\documentclass{article} 
\usepackage{iclr2027_conference,times}

\usepackage{amsmath,amsfonts,bm}

\def\eqref#1{equation~\ref{#1}}

\def\1{\bm{1}}

\DeclareMathAlphabet{\mathsfit}{\encodingdefault}{\sfdefault}{m}{sl}
\SetMathAlphabet{\mathsfit}{bold}{\encodingdefault}{\sfdefault}{bx}{n}

\usepackage{hyperref}
\usepackage{url}
\usepackage{booktabs}
\usepackage{multirow}
\usepackage{graphicx}
\usepackage{array}
\usepackage{amsmath}
\usepackage{amssymb}
\usepackage{placeins}

\newcommand{\UBDTableStyle}{%
  \fontsize{8pt}{8.4pt}\selectfont
  \setlength{\tabcolsep}{2pt}%
  \renewcommand{\arraystretch}{1.08}%
}

\title{UBD-Bench: Benchmarking and Analyzing Unimodal Branch Degradation in Multimodal Perception for Autonomous Driving}

\author{
    \textbf{Ang Liu}\textsuperscript{\rm 1,2}$^{\ast}$,
    \textbf{Xiang Gu}\textsuperscript{\rm 1}$^{\ast}$,
    \textbf{Ruikai Li}\textsuperscript{\rm 1,2}$^{\ast \: \ddagger}$,
    \textbf{Qiyue Wang}\textsuperscript{\rm 1},
    \textbf{Yizhe Fan}\textsuperscript{\rm 1}, \\
    \textbf{Hao Shan}\textsuperscript{\rm 1,2}, 
    \textbf{Ziyang Yan}\textsuperscript{\rm 1,2}, 
    \textbf{Han Jiang}\textsuperscript{\rm 1}$^{\dagger}$,
    \textbf{Zhenning Li}\textsuperscript{\rm 3}, 
    \textbf{Xiaoshuai Hao}\textsuperscript{\rm 2}, \\
    \textbf{Yilong Ren}\textsuperscript{\rm 1}, 
    \textbf{Zhiyong Cui}\textsuperscript{\rm 1},
    \textbf{Haiyang Yu}\textsuperscript{\rm 1}\\
    \textsuperscript{\rm 1} Beihang University \qquad
    \textsuperscript{\rm 2} Xiaomi EV  \qquad
    \textsuperscript{\rm 3} University of Macau \\
    {\small $^{\ast}$Equal contribution \qquad $^{\dagger}$Corresponding author \qquad $^{\ddagger}$Project leader}
}

\iclrfinalcopy
\begin{document}
\setlength{\textfloatsep}{6pt}
\setlength{\floatsep}{6pt}
\setlength{\intextsep}{6pt}
\setlength{\dbltextfloatsep}{6pt}
\setlength{\dblfloatsep}{6pt}

\maketitle

\vspace{-7mm}
\begin{abstract}
\vspace{-3mm}

Multimodal perception improves autonomous-driving performance by combining complementary sensor information. However, stronger multimodal predictions do not necessarily imply stronger representations in each modality-specific branch. Such branch-level weaknesses can remain hidden in conventional output-level evaluation. We introduce UBD-Bench, a systematic benchmark for studying unimodal branch degradation (UBD) in multimodal perception. UBD-Bench uses a Paired Frozen-Branch Probe to compare the task-relevant representation quality of the same modality after independent unimodal training and multimodal joint training under matched decoding conditions. Across four autonomous-driving perception tasks and 10 models, we find that Camera branches consistently degrade after multimodal joint training, whereas LiDAR branches exhibit weaker and more model-dependent changes, revealing a pronounced modality asymmetry. Hierarchical probing further shows that Camera-branch degradation extends from the visual backbone through the Camera-to-BEV pathway and is most severe at the pre-fusion Camera BEV representation. Guided by this diagnosis, we introduce unimodal-to-multimodal distillation at the degraded representation. This intervention consistently improves both Camera-branch Probe performance and full-model performance across all four tasks. These results show that multimodal gains can coexist with degraded unimodal representations, highlighting branch-level representation quality as an important dimension for evaluating and improving multimodal perception.
\end{abstract}

\vspace{-5mm}
\section{Introduction}
\vspace{-2mm}

Accurate and reliable perception of the surrounding environment is fundamental to safe autonomous driving. To this end, camera–LiDAR fusion has become a prevailing paradigm for autonomous-driving perception, as cameras capture dense appearance and semantic cues while LiDAR provides direct and metrically accurate 3D geometry~\citep{caesar2020nuscenes,yan2023cross}. Building on this complementarity, multimodal systems have achieved substantial progress across a broad range of autonomous-driving perception tasks, including 3D object detection~\citep{liu2023bevfusion}, online vectorized map construction~\citep{liao2022maptr}, bird's-eye-view (BEV) map segmentation~\citep{liu2023bevfusion}, and occupancy prediction~\citep{wang2023openoccupancy}. Despite substantial differences in model architecture and prediction target, these systems are commonly evaluated using task-level metrics computed from their final predictions, enabling comparisons of overall perception performance.

\begin{figure}[t]
    \centering
    \includegraphics[width=\linewidth]{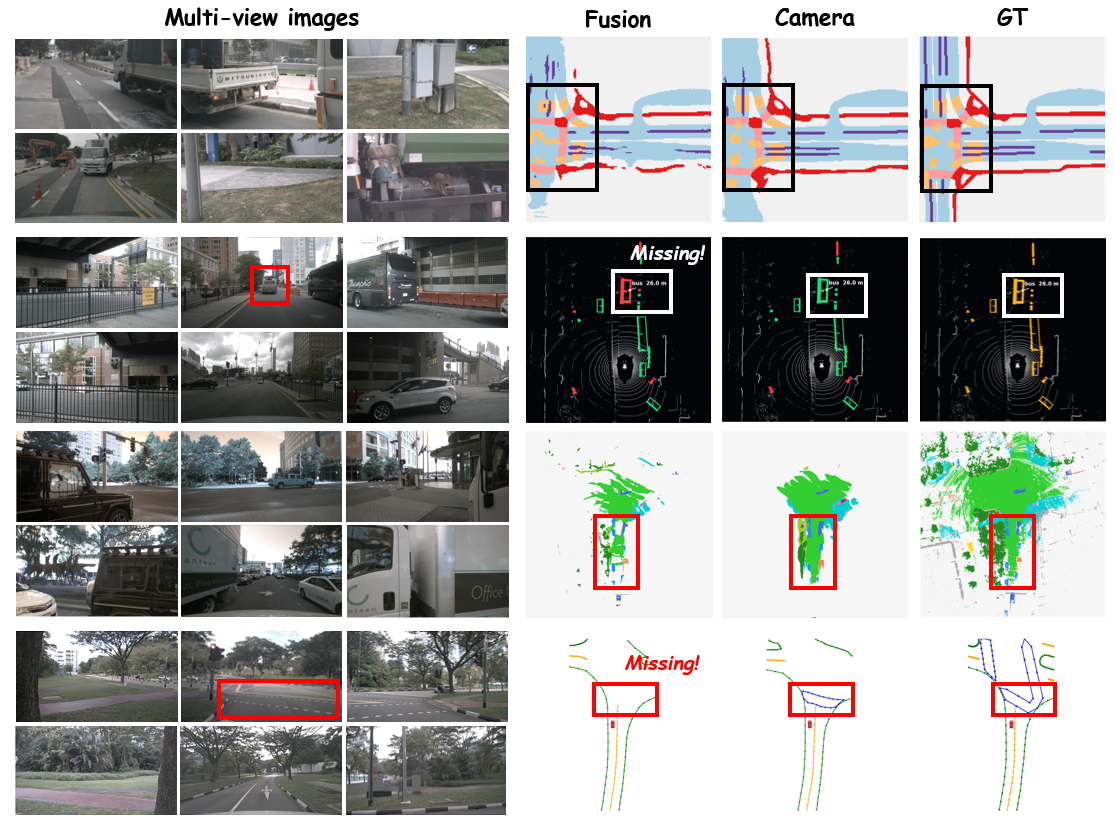}
    \vspace{-8mm}
    \caption{\textbf{Prediction Discrepancies Between Camera--LiDAR and Camera-Only Models.}
Representative examples from 3D object detection, BEV map segmentation, occupancy prediction, and online vectorized map construction show that Camera--LiDAR models produce less accurate predictions than their Camera-only counterparts in the highlighted regions.}
    \label{fig:intro_motivation}
\end{figure}

A natural expectation of multimodal perception is that combining complementary sensor observations should produce more accurate and reliable predictions than relying on either sensor alone. However, Figure~\ref{fig:intro_motivation} shows that this overall advantage does not hold uniformly across scenarios: although the Camera–LiDAR system achieves stronger overall performance, the Camera-only system produces more accurate predictions in certain scenarios.

These examples do not imply that single-sensor perception is generally superior. Instead, they raise a fundamental question: does each modality-specific branch in a jointly trained multimodal model learn a task-relevant representation as effective as that learned by an independently trained unimodal model? Since the two systems differ in both their sensor inputs and processing pipelines, an output-level comparison cannot answer this question or isolate the influence of joint learning on any individual modality. Current evaluations primarily report metrics based on these predictions, providing limited insight into the quality of task-relevant representations learned by individual branches of multimodal models relative to independently trained unimodal models.

To address this gap, we introduce \textbf{UBD-Bench}, a benchmark that uses a \textbf{Paired Frozen-Branch Probe} to assess the task-relevant representational quality of the same modality after independent unimodal training and multimodal joint training. We define \textbf{unimodal branch degradation (UBD)} as lower probe performance for the jointly trained branch than for its independently trained counterpart. UBD-Bench then examines how widely UBD occurs across tasks and architectures, where it arises along modality-specific representation pathways, and whether targeted interventions mitigate it.Our key contributions are threefold:
\vspace{-3mm}

\begin{itemize}
    \setlength{\itemsep}{0.8em}

    \item \textbf{UBD-Bench establishes the first systematic benchmark for unimodal branch degradation in multimodal autonomous-driving perception.}
    Built on the Paired Frozen-Branch Probe, UBD-Bench spans four perception tasks and 10 models, demonstrating the prevalence of Camera-branch degradation across tasks and models, as well as a pronounced modality asymmetry between Camera and LiDAR.
   
\vspace{-3mm} 

    \item \textbf{A hierarchical Probe is developed to localize degradation along the Camera–BEV generation pathway.}
    Cross-task results show that the degradation is already embedded in the branch representation before the task decoder and involves the entire pathway from image encoding to BEV generation, with the most pronounced gap appearing at the pre-fusion Camera BEV representation.

\vspace{-3mm}
    
    \item \textbf{Guided by the hierarchical Probe results, a unimodal-to-multimodal reverse-distillation scheme is designed to intervene in the degraded branch.}
    Transferring knowledge from an independently trained Camera teacher to the corresponding branch of the multimodal model improves both the task-relevant representation quality measured by the branch Probe and full-model performance, providing intervention-based evidence for the degradation diagnosis and demonstrating that targeted representation enhancement can mitigate unimodal branch degradation.
\end{itemize}

\section{Related Work}
\vspace{-2mm}

\noindent\textbf{Multimodal Perception for Autonomous Driving.}
Multimodal perception for autonomous driving combines complementary information to perceive the surrounding environment, with camera–LiDAR fusion serving as a prominent example. Recent research has advanced attention-based interaction between heterogeneous sensor features~\citep{yan2023cross} and shared BEV fusion for 3D detection and map segmentation~\citep{liu2023bevfusion}, with related developments extending multimodal perception to online map construction~\citep{zhang2024gemap,hao2024mapdistill,zhou2024himap} and occupancy prediction~\citep{ming2024occfusion,shi2024streamingflow,li2025occmamba}. These studies span diverse fusion mechanisms and perception tasks. UBD-Bench draws on this diversity to assess whether unimodal branch degradation recurs across tasks and model architectures.

\vspace{-1mm}
\noindent\textbf{Evaluation of Multimodal Perception for Autonomous Driving.}
Evaluation of multimodal perception for autonomous driving examines whether perception models remain accurate and reliable across different tasks, operating conditions, and sensor configurations~\citep{song2024robustness}. Recent benchmarks evaluate natural corruptions in 3D detection~\citep{dong2023benchmarking}, corruption robustness and domain shifts in BEV perception~\citep{xie2025benchmarking}, partial or complete sensor failures in detection and map segmentation~\citep{ge2023metabev}, and combined Camera–LiDAR corruptions in online map construction~\citep{hao2024your}. These evaluations reveal how changes in sensing conditions affect final predictions, but do not directly isolate the effect of multimodal joint training on the task-relevant representation quality of each modality-specific branch under uncorrupted inputs. UBD-Bench complements existing evaluations by using a unified frozen-branch probing protocol to compare representations learned through independent unimodal training and multimodal joint training.

\vspace{-1mm}
\noindent\textbf{Knowledge Distillation for Autonomous-Driving Perception.}
Knowledge distillation in autonomous-driving perception typically transfers task-relevant knowledge from a stronger teacher model to a student model to improve the student's perception performance~\citep{yan2026distillsurvey}. Recent studies have transferred geometric knowledge from LiDAR teachers to Camera detectors~\citep{wang2023distillbev,li2026occdistill,li2023bev-lgkd}, supported knowledge transfer among different sensor configurations through aligned BEV representations~\citep{zhou2023unidistill,zhao2024crkd}, distilled multimodal knowledge into lightweight Camera map models~\citep{hao2024mapdistill,yan2025mapkd}, and used future observations to supervise models operating on current-frame inputs~\citep{li2026amap}. Their improvements show that prediction-, feature-, and relation-level constraints can preserve useful geometric and semantic structure in representations that would otherwise remain comparatively weak. Building on this finding, UBD-Bench transfers knowledge from independently trained unimodal teachers to the corresponding branches during multimodal joint training, and evaluates the intervention through branch Probe performance and fusion performance.

\vspace{-4mm}
\section{Analysis}
\label{sec:analysis}
\vspace{-2mm}

\subsection{Evaluation Settings and Unified Protocol}
\label{sec:protocol}
\vspace{-1mm}

\paragraph{Paired Frozen-Branch Probe.}
A typical Camera--LiDAR perception model comprises a Camera--BEV generation pathway, a LiDAR--BEV generation pathway, a cross-modal fusion module, and a task decoder~\citep{liu2023bevfusion,wang2023openoccupancy}. The final task metric reflects the combined effects of the fused representation and the task decoder and therefore does not directly measure the task-relevant representation quality of either modality. To enable branch-level evaluation, UBD-Bench introduces the \textbf{Paired Frozen-Branch Probe}. The Probe compares representations of the same modality learned through independent unimodal training and multimodal joint training under matched decoding conditions.

\begin{figure}[t]
    \centering
    \includegraphics[width=\linewidth]{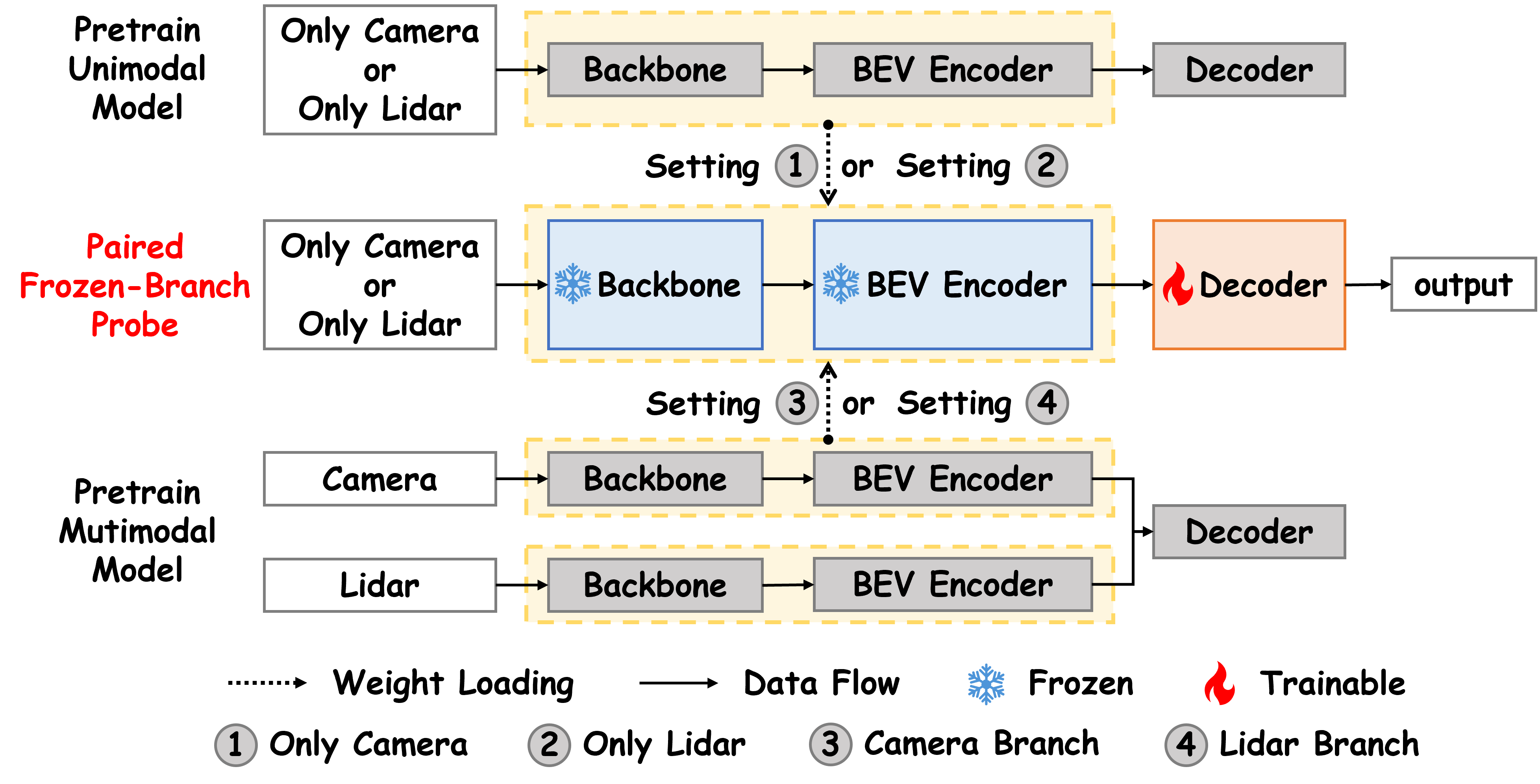}
    \vspace{-8mm}
    \caption{\textbf{Paired Frozen-Branch Probe for Camera and LiDAR representations.}}
    \label{fig:paired_probe}
\end{figure}

For a given probing position, the network from the raw sensor input to that position is defined as the representation extractor $g$. A paired Probe instantiates two such extractors. One extractor loads its weights from a fully trained unimodal model, whereas the other loads the corresponding branch weights from a fully trained multimodal model. Throughout the Probe, both extractors remain frozen and are each connected to a separate task decoder, with the two decoders having the same architecture and identical initial weights. Only the decoders are trained, using identical data splits, optimization settings, and training budgets. The resulting Probe models are evaluated under the same procedure. Consequently, the source of the frozen representation extractor is the only variable within each paired comparison. Figure~\ref{fig:paired_probe} illustrates the weight loading and probing procedure for the Camera and LiDAR branches.

\vspace{-2mm}
\paragraph{Probe Performance.}
Let $q$ denote the task decoder attached to the frozen representation extractor $g$. With $g$ fixed, $q_g^{*}$ denotes the decoder obtained by training $q$ under the Probe setting defined above. The Probe performance of $g$ is defined as
\begin{equation}
P(g)=S\!\left(q_g^{*}\circ g\right),
\label{eq:probe-performance}
\end{equation}
where $S(\cdot)$ is the task-specific metric on a fixed validation set, and $\circ$ denotes function composition. Under a fixed decoder architecture and training setting, $P(g)$ measures how effectively the representation from $g$ supports the target task, and serves as our operational measure of task-relevant representation quality.

\noindent\textbf{Probe Configurations.}
To identify both the training source and the representation being evaluated, each Probe configuration is denoted by $A2B$. The prefix $A$ specifies how the source model was trained: $C$, $L$, and $F$ denote Camera-only, LiDAR-only, and Camera--LiDAR joint training, respectively. The suffix $B$ specifies the representation type, with the same symbols denoting Camera, LiDAR, and fused Camera--LiDAR representations. Let $g_B^A$ denote the type-$B$ representation extractor from a model trained under setting $A$, and let $P(g_B^A)$ denote its Probe performance. For example, C2C evaluates the Camera representation from an independently trained Camera-only model and provides the reference score $P(g_C^C)$, whereas F2C evaluates the Camera representation from the corresponding branch of a jointly trained multimodal model, yielding $P(g_C^F)$.

\vspace{-2mm}

\paragraph{Paired Comparison and Degradation Measure.}
C2C and F2C form a paired evaluation of Camera representations, while L2L and F2L form the corresponding evaluation of LiDAR representations. For modality $v\in\{C,L\}$, we define the branch performance difference as
\begin{equation}
\Delta_v=P\!\left(g_v^F\right)-P\!\left(g_v^v\right),\qquad v\in\{C,L\}.
\label{eq:branch-difference}
\end{equation}
For the Camera and LiDAR branches, the corresponding differences are
\begin{equation}
\Delta_C=P\!\left(g_C^F\right)-P\!\left(g_C^C\right),\qquad
\Delta_L=P\!\left(g_L^F\right)-P\!\left(g_L^L\right).
\label{eq:camera-lidar-differences}
\end{equation}

\begin{figure}[t]
    \centering
    \includegraphics[width=\linewidth]{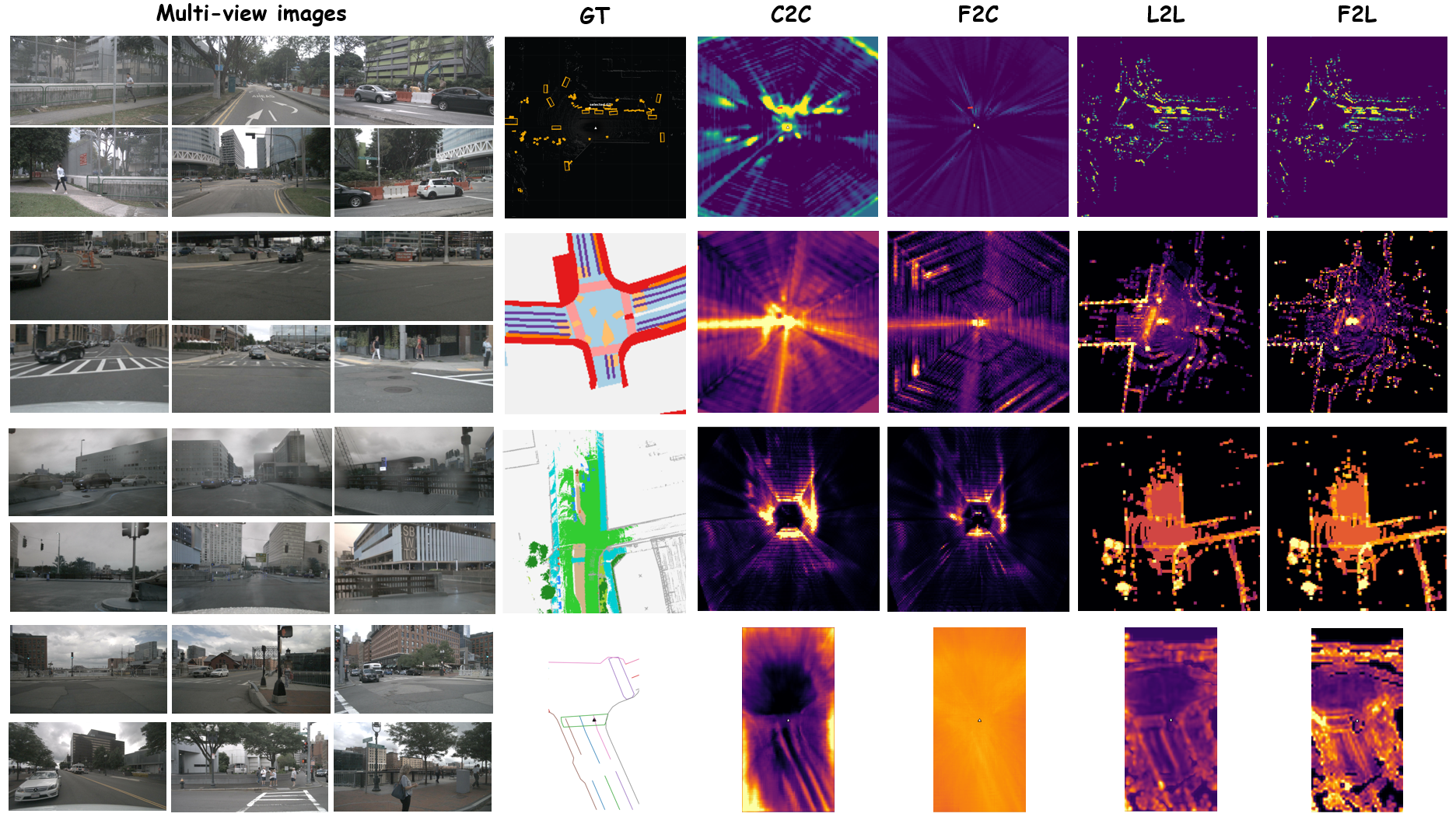}
    \vspace{-8mm}
    \caption{\textbf{Qualitative comparison of modality-specific pre-fusion BEV representations across four perception tasks.} The independently trained Camera and LiDAR representations are compared with the corresponding branches learned through multimodal joint training.}
    \label{fig:cross_task_visualization}
\end{figure}

A negative $\Delta_v$ indicates that the modality-specific branch in the multimodal model achieves lower Probe performance than its independently trained unimodal counterpart. We define this outcome as unimodal branch degradation.

\vspace{-2mm}
\subsection{Unimodal Branch Degradation across Tasks and Models}
\label{sec:cross-task}
\vspace{-1mm}

\paragraph{Tasks and Models.}
UBD-Bench evaluates unimodal branch degradation across four autonomous-driving perception tasks: 3D object detection, online vectorized map construction, BEV map segmentation, and occupancy prediction. These tasks differ in their prediction targets, spatial representations, and task-decoder architectures. Within each task, the selected models further cover different Camera--LiDAR fusion architectures. Specifically, the 3D object detection evaluation includes BEVFusion~\citep{liu2023bevfusion} and CMT~\citep{yan2023cross}. Online vectorized map construction includes MapTR~\citep{liao2022maptr}, HIMap~\citep{zhou2024himap}, MapTRv2~\citep{liao2025maptrv2}, and MapQR~\citep{liu2024leveraging}. BEV map segmentation includes BEVFusion~\citep{liu2023bevfusion}, MapPrior~\citep{zhu2023mapprior}, and DifFUSER~\citep{le2024diffusion}. Occupancy prediction includes CONet~\citep{wang2023openoccupancy}. This evaluation scope enables UBD-Bench to assess whether unimodal branch degradation is prevalent across perception tasks and model architectures.

\vspace{-2mm}

\paragraph{Branch-Level Probe Settings.}
This section employs the Branch-Level Probe, with the probing position fixed at the pre-fusion BEV representation of each modality branch. All representation extraction modules from the raw sensor input to this position remain frozen. The paired comparisons for each model follow the unified protocol in Section~\ref{sec:protocol}. Decoder architectures and training budgets are determined for the corresponding task and matched within each paired comparison. Evaluation uses the metrics adopted in the official implementations of the respective baselines. Specifically, 3D object detection is evaluated using the official nuScenes mAP and NDS~\citep{caesar2020nuscenes}, online vectorized map construction using mAP~\citep{liao2022maptr}, BEV map segmentation using mean IoU over six map categories~\citep{liu2023bevfusion}, and occupancy prediction using IoU and fine-grained semantic mIoU~\citep{wang2023openoccupancy}.

In addition to the branch-level Probe results, we report the end-to-end performance of the original Camera-only, LiDAR-only, and multimodal models, denoted by Full C, Full L, and Full F, respectively. These results characterize task performance under the models' original sensor configurations and inference pipelines. The fused-representation Probe result, F2F, is also reported to compare performance obtained with a frozen fused representation and a newly trained decoder against the original multimodal model's end-to-end performance.

\begin{table}[t]
\centering
\caption{\textbf{Cross-task and cross-model evaluation of unimodal branch degradation.}}
\vspace{2mm}
\label{tab:cross_task_degradation}

\UBDTableStyle
\begin{tabular}{@{}llccccccccc@{}}
\toprule
\multirow{2}{*}{Task}
& \multirow{2}{*}{Method}
& \multirow{2}{*}{Metric}
& \multicolumn{3}{c}{End-to-end}
& \multicolumn{2}{c}{Camera Probe}
& \multicolumn{2}{c}{LiDAR Probe}
& Fusion Probe \\
\cmidrule(lr){4-6}
\cmidrule(lr){7-8}
\cmidrule(lr){9-10}
\cmidrule(l){11-11}
& & & Full C & Full L & Full F & C2C & F2C & L2L & F2L & F2F \\

\midrule
\multirow{4}{*}{3D Det.}
& \multirow{2}{*}{BEVFusion}
& mAP$\uparrow$ & 35.55 & 64.68 & 68.55 & 33.47 & 11.58 (\textbf{-21.89}) & 47.69 & 47.70 (+0.01) & 62.56 \\
& & NDS$\uparrow$ & 41.21 & 69.22 & 71.41 & 36.49 & 19.02 (\textbf{-17.47}) & 53.01 & 53.47 (+0.46) & 61.31 \\
\cmidrule(lr){2-11}
& \multirow{2}{*}{CMT}
& mAP$\uparrow$ & 40.59 & 63.21 & 70.30 & 36.43 & 10.50 (\textbf{-25.93}) & 61.64 & 61.88 (+0.24) & 53.24 \\
& & NDS$\uparrow$ & 46.04 & 68.27 & 72.94 & 41.94 & 25.88 (\textbf{-16.06}) & 67.23 & 67.78 (+0.55) & 60.06 \\

\midrule
\multirow{4}{*}{Vec. Map}
& MapTR   & mAP$\uparrow$ & 48.57 & 54.26 & 63.14 & 51.09 & 40.42 (\textbf{-10.67}) & 53.62 & 55.88 (+2.26) & 62.92 \\
& HIMap   & mAP$\uparrow$ & 65.02 & 65.32 & 73.57 & 64.41 & 57.17 (\textbf{-7.24}) & 65.08 & 64.37 (\textbf{-0.71}) & 72.32 \\
& MapTRv2 & mAP$\uparrow$ & 61.15 & 59.49 & 72.28 & 60.93 & 53.65 (\textbf{-7.28}) & 58.53 & 61.86 (+3.33) & 70.87 \\
& MapQR   & mAP$\uparrow$ & 66.15 & 61.85 & 72.56 & 66.56 & 59.62 (\textbf{-6.94}) & 62.28 & 62.90 (+0.62) & 73.72 \\

\midrule
\multirow{3}{*}{BEV Seg.}
& BEVFusion & mIoU$\uparrow$ & 57.09 & 48.58 & 62.94 & 56.44 & 51.20 (\textbf{-5.24}) & 47.45 & 45.05 (\textbf{-2.40}) & 61.48 \\
& MapPrior  & mIoU$\uparrow$ & 56.76 & 52.41 & 63.15 & 56.56 & 51.34 (\textbf{-5.22}) & 47.69 & 45.67 (\textbf{-2.02}) & 61.73 \\
& DifFUSER  & mIoU$\uparrow$ & 53.74 & 46.28 & 63.51 & 55.26 & 45.11 (\textbf{-10.15}) & 45.97 & 46.91 (+0.94) & 64.83 \\

\midrule
\multirow{2}{*}{Occ.}
& \multirow{2}{*}{CONet}
& IoU$\uparrow$ & 21.50 & 30.00 & 26.50 & 20.80 & 20.10 (\textbf{-0.70}) & 30.50 & 30.50 (+0.00) & 28.90 \\
& & $\mathrm{mIoU}_{\mathrm{fine}}\uparrow$
& 13.60 & 15.90 & 20.50 & 13.10 & 12.00 (\textbf{-1.10}) & 16.10 & 16.20 (+0.10) & 20.60 \\

\bottomrule
\end{tabular}

\vspace{2pt}
\begin{minipage}{0.98\textwidth}
\scriptsize
\textbullet\; Values in parentheses in the F2C and F2L columns denote $\Delta_C$ and $\Delta_L$, respectively. \\[-1pt]
\textbullet\; Negative differences indicate branch degradation and are highlighted in bold.
\end{minipage}
\end{table}

\begin{figure}[t]
    \centering
    \includegraphics[width=1.0\linewidth]{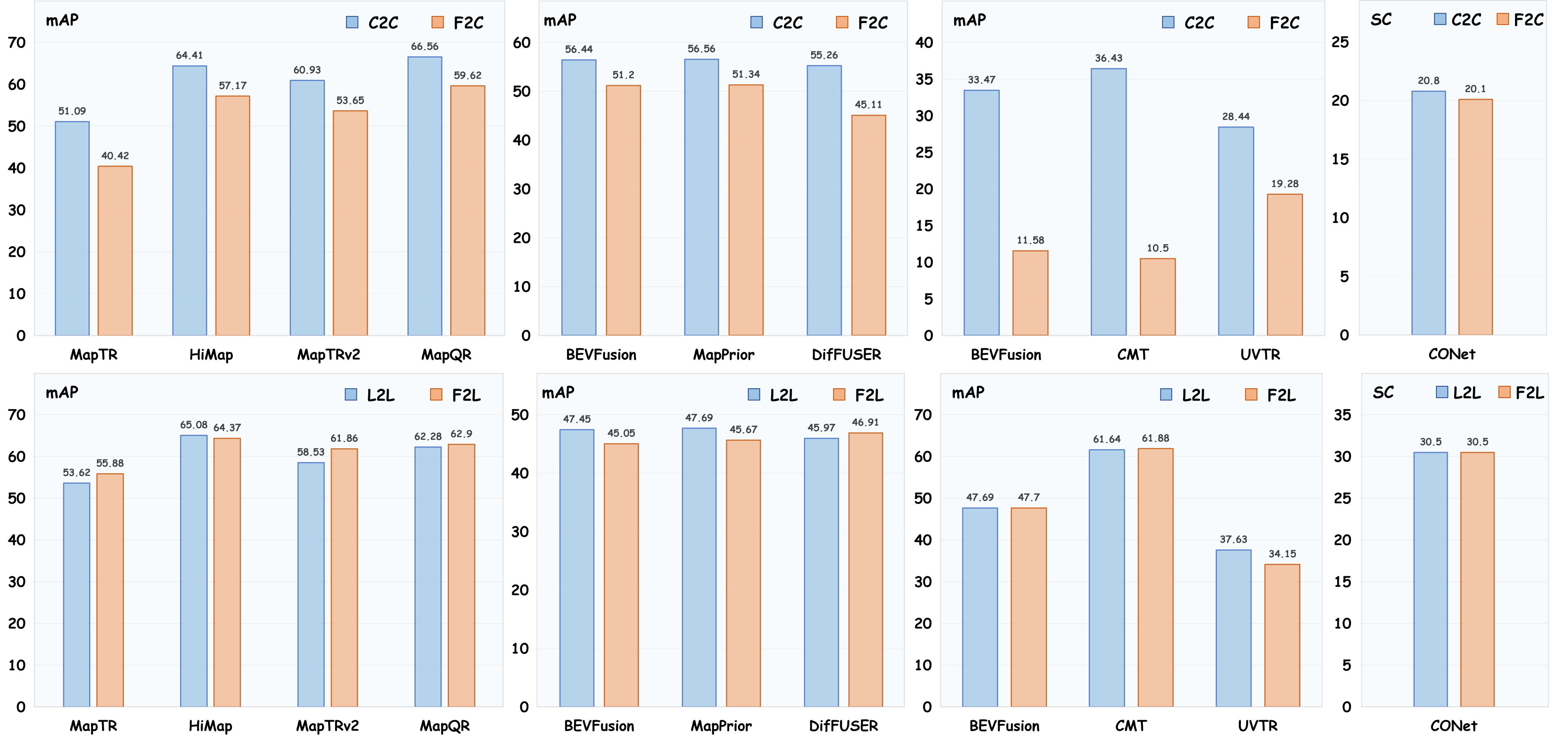}
    \vspace{-5mm}
    \caption{\textbf{Cross-task comparison of Camera- and LiDAR-branch Probe performance.} The top row shows consistent Camera-branch degradation across tasks and models, whereas the bottom row shows more model-dependent changes in the LiDAR branch. Blue and orange bars denote independently trained unimodal branches and
  the corresponding multimodal-model branches, respectively.}
    \label{fig:zzt}
\end{figure}

\vspace{-2mm}

\paragraph{Branch-Level Probe Results.}
As shown in Table~\ref{tab:cross_task_degradation}, \textbf{Camera-branch degradation occurs consistently across tasks and models.} All reported Camera-branch comparisons yield $\Delta_C<0$, indicating that Probe performance under F2C is lower than under C2C. For example, CMT's 3D detection mAP decreases from 36.43 under C2C to 10.50 under F2C, a drop of 25.93 percentage points. MapTR's vectorized mapping mAP decreases by 10.67 percentage points. Despite differences in prediction targets and evaluation metrics, the jointly trained Camera branches consistently achieve lower Probe performance than their independently trained counterparts. This pattern indicates that Camera-branch degradation is not confined to a particular task or model architecture within the evaluated setting.

\noindent\textbf{The LiDAR branch exhibits greater model dependence, revealing an asymmetric pattern of degradation across modalities.}
For CMT, Camera-branch mAP decreases by 25.93 percentage points, whereas LiDAR-branch mAP remains nearly unchanged, increasing from 61.64 to 61.88. Other models exhibit both improvements and declines in LiDAR-branch performance. For example, mAP of MapTRv2 increases by 3.33 percentage points, while BEVFusion's BEV map segmentation mIoU decreases by 2.40 percentage points. These results indicate that the changes in branch representations associated with multimodal joint training differ across modalities. Camera branches consistently exhibit degradation within the current evaluation scope, whereas LiDAR-branch outcomes vary across models and tasks.


\vspace{-4mm}

\paragraph{Qualitative Analysis of Modality-Specific Representations.}
Figure~\ref{fig:cross_task_visualization} qualitatively compares the
pre-fusion BEV representations learned through independent unimodal training
with those learned by the corresponding branches through multimodal joint
training. Across the representative models from all four tasks, the
Fusion-Camera representations exhibit noticeable changes in spatial
organization relative to their Camera-only counterparts. The differences are
particularly pronounced in 3D object detection and online vectorized map
construction, where the Fusion-Camera activations become less spatially
distinctive. In comparison, the change is more subtle for occupancy prediction,
consistent with its smaller Camera-branch performance difference. The
Fusion-LiDAR representations, however, largely preserve the spatial patterns
observed in their LiDAR-only counterparts and do not exhibit a consistent loss
of structure across tasks. These qualitative observations support the Probe
results: Camera-branch degradation occurs consistently across tasks, although
its magnitude varies, whereas the effect of multimodal joint training on the
LiDAR branch depends more strongly on the task and model.

\FloatBarrier
\vspace{-1mm}
\subsection{Hierarchical Probe and Degradation Localization}
\label{sec:hierarchical_probe}
\vspace{-1mm}
The branch-level evaluation in Section~\ref{sec:cross-task} shows that the pre-fusion Camera BEV representations learned through multimodal joint training exhibit degradation. However, only evaluating the branch output cannot determine whether this gap is already present in earlier intermediate representations. To address this question, the Hierarchical Probe progressively moves the probing position upstream along the Camera--BEV generation pathway and retrains all downstream modules to assess representation gaps at different processing stages.

\paragraph{Hierarchical Probe Settings.}
The Camera--BEV generation pathway comprises a visual backbone, an image neck, and a BEV encoder, where the BEV encoder performs view transformation from image features to BEV representations. Five probing configurations, P0--P4, are defined along this pathway. P0 follows the branch-level setting in Section~\ref{sec:cross-task}, freezing the entire Camera--BEV generation pathway and retraining only the task decoder. P1 moves the probing position to the input of the BEV encoder, keeping the visual backbone and image neck frozen while retraining the BEV encoder and task decoder. P2 moves the probing position further upstream to the visual-backbone output, freezing only the backbone and retraining the image neck, BEV encoder, and task decoder. P3 and P4 place the probing position at the inputs to the final one and two backbone stages, respectively. All preceding backbone modules remain frozen, while the corresponding backbone stages and all subsequent modules are retrained.

At each probing position, C2C and F2C follow the paired protocol in Section~\ref{sec:protocol}, with all downstream modules trained from scratch. One representative model is selected for each task: MapTR for online vectorized map construction, BEVFusion for BEV map segmentation and 3D object detection, and CONet for occupancy prediction. Each task retains the training budget and evaluation metrics used in Section~\ref{sec:cross-task}.

\begin{table*}[htbp]
\centering
\caption{\textbf{Hierarchical probing along the Camera--BEV generation pathway.}}
\label{tab:hierarchical_probe}

\vspace{6pt}

\fontsize{8pt}{9.6pt}\selectfont
\setlength{\tabcolsep}{7pt}
\renewcommand{\arraystretch}{1.08}

\begin{minipage}[t]{0.48\textwidth}
\centering
{\fontsize{9pt}{10.8pt}\selectfont\textbf{Vec.\ Map / MapTR}}

\vspace{3pt}

\begin{tabular}{@{\hspace{12pt}}llcc@{\hspace{12pt}}}
\toprule
Metric & Pos. & C2C & F2C ($\Delta_C$) \\
\midrule
\multirow{5}{*}{mAP$\uparrow$}
& P0 & 51.09 & 40.42 (\textbf{-10.67}) \\
& P1 & 51.58 & 43.80 (\textbf{-7.78}) \\
& P2 & 49.84 & 45.73 (\textbf{-4.11}) \\
& P3 & 45.07 & 42.49 (\textbf{-2.58}) \\
& P4 & 26.09 & 26.49 (+0.40) \\
\bottomrule
\end{tabular}
\end{minipage}
\hfill
\begin{minipage}[t]{0.48\textwidth}
\centering
{\fontsize{9pt}{10.8pt}\selectfont\textbf{BEV Seg.\ / BEVFusion}}

\vspace{3pt}

\begin{tabular}{@{\hspace{12pt}}llcc@{\hspace{12pt}}}
\toprule
Metric & Pos. & C2C & F2C ($\Delta_C$) \\
\midrule
\multirow{5}{*}{mIoU$\uparrow$}
& P0 & 56.44 & 51.20 (\textbf{-5.24}) \\
& P1 & 56.46 & 52.04 (\textbf{-4.42}) \\
& P2 & 55.78 & 53.79 (\textbf{-1.99}) \\
& P3 & 55.22 & 53.08 (\textbf{-2.14}) \\
& P4 & 52.38 & 51.02 (\textbf{-1.36}) \\
\bottomrule
\end{tabular}
\end{minipage}

\vspace{9pt}

\begin{minipage}[t]{0.48\textwidth}
\centering
{\fontsize{9pt}{10.8pt}\selectfont\textbf{3D Det.\ / BEVFusion}}

\vspace{3pt}

\begin{tabular}{@{\hspace{12pt}}llcc@{\hspace{12pt}}}
\toprule
Metric & Pos. & C2C & F2C ($\Delta_C$) \\
\midrule
\multirow{5}{*}{mAP$\uparrow$}
& P0 & 33.47 & 11.58 (\textbf{-21.89}) \\
& P1 & 34.43 & 15.11 (\textbf{-19.32}) \\
& P2 & 33.39 & 23.84 (\textbf{-9.55}) \\
& P3 & 33.12 & 26.67 (\textbf{-6.45}) \\
& P4 & 29.22 & 27.51 (\textbf{-1.71}) \\
\midrule
\multirow{5}{*}{NDS$\uparrow$}
& P0 & 36.49 & 19.02 (\textbf{-17.47}) \\
& P1 & 37.88 & 22.04 (\textbf{-15.84}) \\
& P2 & 37.62 & 29.37 (\textbf{-8.25}) \\
& P3 & 37.95 & 32.29 (\textbf{-5.66}) \\
& P4 & 33.64 & 32.63 (\textbf{-1.01}) \\
\bottomrule
\end{tabular}
\end{minipage}
\hfill
\begin{minipage}[t]{0.48\textwidth}
\centering
{\fontsize{9pt}{10.8pt}\selectfont\textbf{Occ.\ / CONet}}

\vspace{3pt}

\begin{tabular}{@{\hspace{12pt}}llcc@{\hspace{12pt}}}
\toprule
Metric & Pos. & C2C & F2C ($\Delta_C$) \\
\midrule
\multirow{5}{*}{IoU$\uparrow$}
& P0 & 20.80 & 20.10 (\textbf{-0.70}) \\
& P1 & 21.50 & 21.30 (\textbf{-0.20}) \\
& P2 & 21.30 & 20.10 (\textbf{-1.20}) \\
& P3 & 20.90 & 20.50 (\textbf{-0.40}) \\
& P4 & 20.10 & 19.50 (\textbf{-0.60}) \\
\midrule
\multirow{5}{*}{$\mathrm{mIoU}_{\mathrm{fine}}\uparrow$}
& P0 & 13.10 & 12.00 (\textbf{-1.10}) \\
& P1 & 13.60 & 13.00 (\textbf{-0.60}) \\
& P2 & 13.40 & 12.90 (\textbf{-0.50}) \\
& P3 & 13.50 & 13.20 (\textbf{-0.30}) \\
& P4 & 11.90 & 11.70 (\textbf{-0.20}) \\
\bottomrule
\end{tabular}
\end{minipage}

\vspace{5pt}

\begin{minipage}{0.98\textwidth}
\raggedright
\textbullet\;
Values in parentheses in the F2C column denote $\Delta_C$. \\[-1pt]
\textbullet\;
Negative differences indicate Camera-branch degradation and are highlighted in bold.
\end{minipage}

\end{table*}

\vspace{-2mm}

\paragraph{Hierarchical Probe Results.}
\textbf{Camera-branch degradation extends throughout the Camera--BEV generation pathway.}
Table~\ref{tab:hierarchical_probe} shows $\Delta_C<0$ at P0--P2 for all four tasks. For vectorized mapping, 3D detection, and BEV segmentation, successively retraining the BEV encoder and image neck reduces the degradation while C2C remains relatively stable. In 3D detection, $\Delta_C$ changes from $-21.89$ at P0 to $-19.32$ at P1 and $-9.55$ at P2, whereas C2C remains between 33.39 and 34.43 mAP. These paired changes indicate that both the BEV encoder and image neck contribute to the observed degradation. Occupancy prediction shows a non-monotonic trend, with the IoU difference decreasing from $-0.20$ at P1 to $-1.20$ at P2 while remaining degraded. Because all downstream modules have been retrained at P2, the remaining negative $\Delta_C$ demonstrates degradation in the frozen visual backbone.

\noindent\textbf{The deep backbone also exhibits degradation, whereas P4 reveals a limitation of the Probe.}
At P3, $\Delta_C$ remains negative for every task. For 3D detection, C2C changes only from 33.39 to 33.12 mAP, while $\Delta_C$ changes from $-9.55$ to $-6.45$ after the deep backbone stage is included in retraining. Vectorized mapping and occupancy prediction show the same reduction in degradation, while BEV segmentation remains degraded by 2.14 points, indicating that degradation also occurs in the deep backbone. At P4, however, C2C decreases across all tasks; for MapTR, C2C/F2C falls from 45.07/42.49 at P3 to 26.09/26.49 mAP. This joint decline indicates that splitting the backbone at this depth prevents it from retaining its original feature-extraction capability, making the P4 Probe unreliable. P4 is therefore excluded from the overall hierarchical comparison. Overall, the visual backbone, image neck, and BEV encoder all exhibit degradation, which is most severe at the pre-fusion Camera BEV representation.

\vspace{-1mm}
\subsection{Distillation Intervention and Degradation Mitigation}
\vspace{-1mm}
\label{sec:distillation}

\paragraph{Unimodal-to-Multimodal Distillation.}

The hierarchical probing results in
Section~\ref{sec:hierarchical_probe} show that Camera-branch degradation
occurs in the visual backbone, image neck, and BEV encoder and is most
pronounced overall at the pre-fusion Camera BEV representation.
Motivated by this finding, we introduce targeted unimodal-to-multimodal
distillation at this position to examine whether improving the task-relevant
representation quality of the Camera branch can mitigate the degradation and
improve the task performance of the multimodal model.

Specifically, an independently trained Camera-only model serves as the teacher, while the Camera branch of the multimodal model serves as the student. The teacher and student process the same Camera input, and feature distillation is applied to their pre-fusion Camera BEV representations. Since knowledge is transferred from an independently trained unimodal model to the corresponding branch of a multimodal model, we refer to this procedure as \textbf{unimodal-to-multimodal distillation}.

Given an input image $x$, the frozen teacher and the student Camera branch produce pre-fusion Camera BEV features $F_{\mathrm{cam},T}(x)$ and $F_{\mathrm{cam},S}(x)$, respectively, where the subscripts $T$ and $S$ denote the teacher and student. The distillation loss is defined as
\begin{equation}
\mathcal{L}_{\mathrm{KD}}
=
\operatorname{MSE}
\left(
F_{\mathrm{cam},S}(x),
F_{\mathrm{cam},T}(x)
\right).
\label{eq:distillation-loss}
\end{equation}
The loss is computed over all spatial positions and channels of the Camera BEV feature map. The multimodal model is optimized using
\begin{equation}
\mathcal{L}
=
\mathcal{L}_{\mathrm{task}}
+
\lambda_t \mathcal{L}_{\mathrm{KD}},
\label{eq:overall-distillation-objective}
\end{equation}
where $\mathcal{L}_{\mathrm{task}}$ is the original training objective for task $t$, and $\lambda_t$ controls the strength of the distillation constraint. The teacher remains frozen throughout training and is removed afterward. Therefore, the proposed intervention does not alter the inference architecture or introduce additional computational overhead during inference.

\vspace{-2mm}

\paragraph{Experimental Settings.}
The distillation experiments use the representative models selected for the four tasks in Section~\ref{sec:hierarchical_probe}. BEVFusion is used for 3D object detection and BEV map segmentation, MapTR for online vectorized map construction, and CONet for occupancy prediction. For each task, the multimodal baseline trained without distillation and its distilled counterpart use the same model architecture, parameter initialization, data processing, optimization settings, training budget, and evaluation procedure. The distillation loss is the only additional training constraint.

After training, the full multimodal model is first evaluated using its standard inference pipeline. Its Camera branch is then evaluated using the unified Probe protocol defined in Section~\ref{sec:protocol}. These two evaluations measure the effects of distillation on the task-relevant representation quality of the Camera branch and the final task performance of the multimodal model, respectively.

\begin{table}[t]
\centering
\caption{\textbf{Effects of unimodal-to-multimodal reverse distillation.}}
\label{tab:distillation_intervention}

\vspace{6pt}

\UBDTableStyle

\begin{tabular}{@{}lllrrrr@{}}
\toprule
\multirow{2}{*}{Task}
& \multirow{2}{*}{Method}
& \multirow{2}{*}{Metric}
& \multicolumn{2}{c}{Full multimodal model}
& \multicolumn{2}{c}{Camera-branch Probe} \\
\cmidrule(lr){4-5}
\cmidrule(lr){6-7}
& & & Baseline & +KD ($\Delta$) & F2C & F2C + KD ($\Delta$) \\
\midrule

\multirow{2}{*}{3D Det.}
& \multirow{2}{*}{BEVFusion}
& mAP$\uparrow$
& 68.27 & \textbf{69.25} (+0.98)
& 12.06 & \textbf{32.48} (+20.42) \\
& & NDS$\uparrow$
& 71.29 & \textbf{71.88} (+0.59)
& 19.06 & \textbf{35.12} (+16.06) \\
\midrule

Vec. Map
& MapTR
& mAP$\uparrow$
& 61.97 & \textbf{64.49} (+2.52)
& 40.42 & \textbf{43.54} (+3.12) \\
\midrule

BEV Seg.
& BEVFusion
& mIoU$\uparrow$
& 59.83 & \textbf{60.45} (+0.62)
& 49.30 & \textbf{55.46} (+6.16) \\
\midrule

\multirow{2}{*}{Occ.}
& \multirow{2}{*}{CONet}
& IoU$\uparrow$
& 25.30 & \textbf{29.30} (+4.00)
& 20.10 & \textbf{21.50} (+1.40) \\
& & $\mathrm{mIoU}_{\mathrm{fine}}\uparrow$
& 20.00 & \textbf{20.80} (+0.80)
& 12.00 & \textbf{12.90} (+0.90) \\

\bottomrule
\end{tabular}

\vspace{4pt}

\begin{minipage}{0.98\linewidth}
\raggedright
\textbullet\;
Parentheses denote absolute changes from matched baselines and bold indicates improvements.
\end{minipage}

\end{table}


\paragraph{Distillation Results.}
\textbf{Unimodal-to-multimodal distillation improves the task-relevant representation quality of the Camera branch.}
As shown in Table~\ref{tab:distillation_intervention}, the distilled Camera branch achieves higher Probe performance on every reported metric. For example, its F2C mAP for 3D object detection improves by 20.42 percentage points. The consistent improvements across tasks indicate that representation supervision from an independently trained Camera-only model can effectively mitigate Camera-branch degradation under multimodal joint training.

\noindent\textbf{Improving the Camera-branch representation further benefits the task performance of the full multimodal model.}
All reported end-to-end metrics improve after distillation, including a 4.00-percentage-point improvement in IoU for occupancy prediction. The concurrent improvements in branch-level Probe performance and end-to-end task performance indicate that the diagnosed degradation can be mitigated through targeted representation supervision and that improving the affected Camera representation benefits final multimodal prediction.

\begin{figure}[t]
    \centering
    \includegraphics[width=\linewidth]{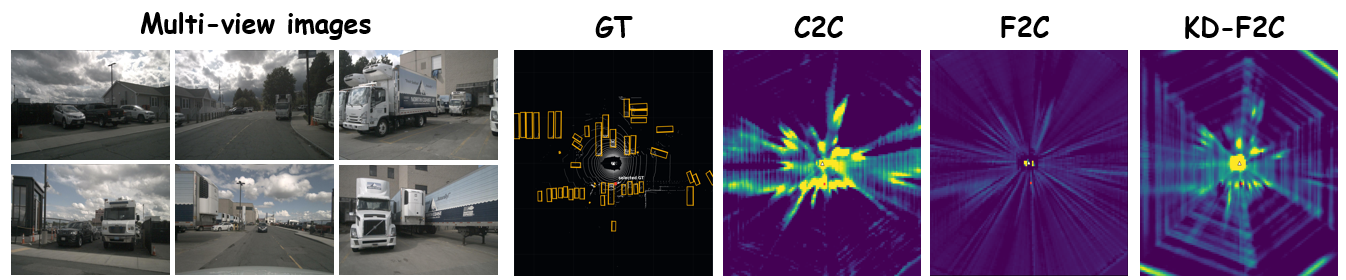}
    \vspace{-8mm}
    \caption{\textbf{Qualitative comparison of pre-fusion Camera BEV features for 3D object detection.} From left to right: ground-truth 3D bounding boxes, the C2C reference feature, the F2C feature without distillation, and the F2C feature after unimodal-to-multimodal distillation. The same feature channel and color scale are used for all activation maps.}
    \label{fig:distillation_visualization}
\end{figure}

\vspace{-4mm}

\paragraph{Qualitative Visualization.}
Figure~\ref{fig:distillation_visualization} provides a representative visualization of the pre-fusion Camera BEV features for 3D object detection. Without distillation, the F2C feature exhibits substantially weaker activations and less distinct spatial structure than the C2C reference. After unimodal-to-multimodal distillation, structured responses become more apparent in the forward region and around the selected object, making the feature qualitatively closer to the C2C reference. This example visually complements the consistent improvements in Camera-branch Probe performance reported in Table~\ref{tab:distillation_intervention}.

\vspace{-3mm}
\section{Conclusion}
\label{sec:conclusion}
\vspace{-2mm}
This work introduces UBD-Bench for systematically evaluating unimodal branch degradation in multimodal autonomous-driving perception. Across four perception tasks and 10 models, we observe prevalent Camera-branch degradation and a clear asymmetry between Camera and LiDAR. Hierarchical probing further shows that the degradation spans the Camera–BEV pathway and is most pronounced at the pre-fusion Camera BEV representation. Distillation at this position consistently improves both Camera-branch Probe performance and end-to-end performance across all four tasks, providing intervention-based evidence for the diagnosis and demonstrating that the degradation can be mitigated. Overall, UBD-Bench provides a unified framework for measuring and improving modality-specific representations under multimodal joint training.

\clearpage



\bibliography{iclr2027_conference}

@inproceedings{caesar2020nuscenes,
  title={nuscenes: A multimodal dataset for autonomous driving},
  author={Caesar, Holger and Bankiti, Varun and Lang, Alex H and Vora, Sourabh and Liong, Venice Erin and Xu, Qiang and Krishnan, Anush and Pan, Yu and Baldan, Giancarlo and Beijbom, Oscar},
  booktitle={2020 IEEE/CVF conference on computer vision and pattern recognition (CVPR)},
  pages={11618--11628},
  year={2020},
  organization={IEEE}
}

@inproceedings{dong2023benchmarking,
  title={Benchmarking robustness of 3D object detection to common corruptions in autonomous driving},
  author={Dong, Yinpeng and Kang, Caixin and Zhang, Jinlai and Zhu, Zijian and Wang, Yikai and Yang, Xiao and Su, Hang and Wei, Xingxing and Zhu, Jun},
  booktitle={2023 IEEE/CVF Conference on Computer Vision and Pattern Recognition (CVPR)},
  pages={1022--1032},
  year={2023},
  organization={IEEE}
}

@inproceedings{ge2023metabev,
  title={Metabev: Solving sensor failures for 3d detection and map segmentation},
  author={Ge, Chongjian and Chen, Junsong and Xie, Enze and Wang, Zhongdao and Hong, Lanqing and Lu, Huchuan and Li, Zhenguo and Luo, Ping},
  booktitle={2023 IEEE/CVF International Conference on Computer Vision (ICCV)},
  pages={8687--8697},
  year={2023},
  organization={IEEE}
}

@inproceedings{hao2024mapdistill,
  title={Mapdistill: Boosting efficient camera-based hd map construction via camera-lidar fusion model distillation},
  author={Hao, Xiaoshuai and Li, Ruikai and Zhang, Hui and Li, Dingzhe and Yin, Rong and Jung, Sangil and Park, Seung-In and Yoo, ByungIn and Zhao, Haimei and Zhang, Jing},
  booktitle={European Conference on Computer Vision},
  pages={166--183},
  year={2024},
  organization={Springer}
}

@article{hao2024your,
  title={Is your hd map constructor reliable under sensor corruptions?},
  author={Hao, Xiaoshuai and Wei, Mengchuan and Yang, Yifan and Zhao, Haimei and Zhang, Hui and Zhou, Yi and Wang, Qiang and Li, Weiming and Kong, Lingdong and Zhang, Jing},
  journal={Advances in Neural Information Processing Systems},
  volume={37},
  pages={22441--22482},
  year={2024}
}

@inproceedings{le2024diffusion,
  title={Diffusion model for robust multi-sensor fusion in 3d object detection and bev segmentation},
  author={Le, Duy-Tho and Shi, Hengcan and Cai, Jianfei and Rezatofighi, Hamid},
  booktitle={European conference on computer vision},
  pages={232--249},
  year={2024},
  organization={Springer}
}

@inproceedings{li2026amap,
  title={Amap: Distilling future priors for ahead-aware online hd map construction},
  author={Li, Ruikai and Li, Xinrun and Xie, Mengwei and Shan, Hao and Qiu, Shoumeng and Chang, Xinyuan and Fan, Yizhe and Xiong, Feng and Jiang, Han and Ren, Yilong and others},
  booktitle={Proceedings of the IEEE/CVF Conference on Computer Vision and Pattern Recognition},
  pages={24906--24917},
  year={2026}
}

@article{liao2022maptr,
  title={Maptr: Structured modeling and learning for online vectorized hd map construction},
  author={Liao, Bencheng and Chen, Shaoyu and Wang, Xinggang and Cheng, Tianheng and Zhang, Qian and Liu, Wenyu and Huang, Chang},
  journal={arXiv preprint arXiv:2208.14437},
  year={2022}
}

@article{liao2025maptrv2,
  title={Maptrv2: An end-to-end framework for online vectorized hd map construction},
  author={Liao, Bencheng and Chen, Shaoyu and Zhang, Yunchi and Jiang, Bo and Zhang, Qian and Liu, Wenyu and Huang, Chang and Wang, Xinggang},
  journal={International Journal of Computer Vision},
  volume={133},
  number={3},
  pages={1352--1374},
  year={2025},
  publisher={Springer}
}

@inproceedings{liu2023bevfusion,
  title={Bevfusion: Multi-task multi-sensor fusion with unified bird's-eye view representation},
  author={Liu, Zhijian and Tang, Haotian and Amini, Alexander and Yang, Xinyu and Mao, Huizi and Rus, Daniela L and Han, Song},
  booktitle={2023 IEEE international conference on robotics and automation (ICRA)},
  pages={2774--2781},
  year={2023},
  organization={ieee}
}

@inproceedings{liu2024leveraging,
  title={Leveraging enhanced queries of point sets for vectorized map construction},
  author={Liu, Zihao and Zhang, Xiaoyu and Liu, Guangwei and Zhao, Ji and Xu, Ningyi},
  booktitle={European Conference on Computer Vision},
  pages={461--477},
  year={2024},
  organization={Springer}
}

@inproceedings{shi2024streamingflow,
  title={StreamingFlow: Streaming occupancy forecasting with asynchronous multi-modal data streams via neural ordinary differential equation},
  author={Shi, Yining and Jiang, Kun and Wang, Ke and Li, Jiusi and Wang, Yunlong and Yang, Mengmeng and Yang, Diange},
  booktitle={2024 IEEE/CVF Conference on Computer Vision and Pattern Recognition (CVPR)},
  pages={14833--14842},
  year={2024},
  organization={IEEE}
}

@inproceedings{wang2023openoccupancy,
  title={Openoccupancy: A large scale benchmark for surrounding semantic occupancy perception},
  author={Wang, Xiaofeng and Zhu, Zheng and Xu, Wenbo and Zhang, Yunpeng and Wei, Yi and Chi, Xu and Ye, Yun and Du, Dalong and Lu, Jiwen and Wang, Xingang},
  booktitle={2023 IEEE/CVF International Conference on Computer Vision (ICCV)},
  pages={17804--17813},
  year={2023},
  organization={IEEE}
}

@inproceedings{wang2023distillbev,
  title={Distillbev: Boosting multi-camera 3d object detection with cross-modal knowledge distillation},
  author={Wang, Zeyu and Li, Dingwen and Luo, Chenxu and Xie, Cihang and Yang, Xiaodong},
  booktitle={2023 IEEE/CVF International Conference on Computer Vision (ICCV)},
  pages={8603--8612},
  year={2023},
  organization={IEEE}
}

@article{xie2025benchmarking,
  title={Benchmarking and improving bird’s eye view perception robustness in autonomous driving},
  author={Xie, Shaoyuan and Kong, Lingdong and Zhang, Wenwei and Ren, Jiawei and Pan, Liang and Chen, Kai and Liu, Ziwei},
  journal={IEEE Transactions on Pattern Analysis and Machine Intelligence},
  volume={47},
  number={5},
  pages={3878--3894},
  year={2025},
  publisher={IEEE}
}

@inproceedings{yan2023cross,
  title={Cross modal transformer: Towards fast and robust 3d object detection},
  author={Yan, Junjie and Liu, Yingfei and Sun, Jianjian and Jia, Fan and Li, Shuailin and Wang, Tiancai and Zhang, Xiangyu},
  booktitle={2023 IEEE/CVF International Conference on Computer Vision (ICCV)},
  pages={18222--18232},
  year={2023},
  organization={IEEE}
}

@inproceedings{zhou2023unidistill,
  title={UniDistill: A universal cross-modality knowledge distillation framework for 3D object detection in bird's-eye view},
  author={Zhou, Shengchao and Liu, Weizhou and Hu, Chen and Zhou, Shuchang and Ma, Chao},
  booktitle={2023 IEEE/CVF Conference on Computer Vision and Pattern Recognition (CVPR)},
  pages={5116--5125},
  year={2023},
  organization={IEEE}
}

@inproceedings{zhou2024himap,
  title={Himap: Hybrid representation learning for end-to-end vectorized hd map construction},
  author={Zhou, Yi and Zhang, Hui and Yu, Jiaqian and Yang, Yifan and Jung, Sangil and Park, Seung-In and Yoo, ByungIn},
  booktitle={2024 IEEE/CVF Conference on Computer Vision and Pattern Recognition (CVPR)},
  pages={15396--15406},
  year={2024},
  organization={IEEE}
}

@article{li2026occdistill,
  title={OccDistill: Distilling Multi-modal Knowledge into Camera-based 3D Occupancy Networks for Autonomous Driving},
  author={Li, Ruikai and Cui, Zhiyong and Shan, Hao and Jiang, Han and Ren, Yilong and Wen, Sijia and Liu, Zhanwen and Yu, Haiyang},
  journal={IEEE Transactions on Multimedia},
  year={2026},
  publisher={IEEE}
}

@inproceedings{zhu2023mapprior,
  title={Mapprior: Bird’s-eye view map layout estimation with generative models},
  author={Zhu, Xiyue and Zyrianov, Vlas and Liu, Zhijian and Wang, Shenlong},
  booktitle={2023 IEEE/CVF International Conference on Computer Vision (ICCV)},
  pages={8194--8205},
  year={2023},
  organization={IEEE}
}

@inproceedings{li2025occmamba,
  title={Occmamba: Semantic occupancy prediction with state space models},
  author={Li, Heng and Hou, Yuenan and Xing, Xiaohan and Ma, Yuexin and Sun, Xiao and Zhang, Yanyong},
  booktitle={2025 IEEE/CVF Conference on Computer Vision and Pattern Recognition (CVPR)},
  pages={11949--11959},
  year={2025},
  organization={IEEE}
}

@article{ming2024occfusion,
  title={Occfusion: Multi-sensor fusion framework for 3d semantic occupancy prediction},
  author={Ming, Zhenxing and Berrio, Julie Stephany and Shan, Mao and Worrall, Stewart},
  journal={IEEE Transactions on Intelligent Vehicles},
  year={2024},
  publisher={IEEE}
}

@inproceedings{zhang2024gemap,
  title={Online vectorized hd map construction using geometry},
  author={Zhang, Zhixin and Zhang, Yiyuan and Ding, Xiaohan and Jin, Fusheng and Yue, Xiangyu},
  booktitle={European Conference on Computer Vision},
  pages={73--90},
  year={2024},
  organization={Springer}
}

@article{li2023bev-lgkd,
  title={Bev-lgkd: A unified lidar-guided knowledge distillation framework for multi-view bev 3d object detection},
  author={Li, Jianing and Lu, Ming and Liu, Jiaming and Guo, Yandong and Du, Yuan and Du, Li and Zhang, Shanghang},
  journal={IEEE Transactions on Intelligent Vehicles},
  volume={9},
  number={1},
  pages={2489--2498},
  year={2023},
  publisher={IEEE}
}

@inproceedings{zhao2024crkd,
  title={Crkd: Enhanced camera-radar object detection with cross-modality knowledge distillation},
  author={Zhao, Lingjun and Song, Jingyu and Skinner, Katherine A},
  booktitle={2024 IEEE/CVF Conference on Computer Vision and Pattern Recognition (CVPR)},
  pages={15470--15480},
  year={2024},
  organization={IEEE}
}

@article{yan2026distillsurvey,
  title={3D object detection and knowledge distillation in autonomous driving: A survey},
  author={Yan, Weiqing and Bu, Changhong and Zhang, Yuanyang and Yue, Guanghui and Zhou, Wei and Zhu, Xiatian and Han, Jungong},
  journal={Information Fusion},
  pages={104706},
  year={2026},
  publisher={Elsevier}
}

@article{song2024robustness,
  title={Robustness-aware 3d object detection in autonomous driving: A review and outlook},
  author={Song, Ziying and Liu, Lin and Jia, Feiyang and Luo, Yadan and Jia, Caiyan and Zhang, Guoxin and Yang, Lei and Wang, Li},
  journal={IEEE Transactions on Intelligent Transportation Systems},
  volume={25},
  number={11},
  pages={15407--15436},
  year={2024},
  publisher={IEEE}
}

@article{yan2025mapkd,
  title={MapKD: Unlocking Prior Knowledge with Cross-Modal Distillation for Efficient Online HD Map Construction},
  author={Yan, Ziyang and Li, Ruikai and Cui, Zhiyong and Li, Bohan and Jiang, Han and Ren, Yilong and Li, Aoyong and Li, Zhenning and Wen, Sijia and Yu, Haiyang},
  journal={arXiv preprint arXiv:2508.15653},
  year={2025}
}
\bibliographystyle{iclr2027_conference}

\end{document}